\documentclass[letterpaper, 10 pt, conference]{ieeeconf}  

\IEEEoverridecommandlockouts                              

\usepackage{xcolor}
\usepackage{amsmath}
\usepackage{amssymb}
\usepackage{mathtools}
\let\labelindent\relax                                    
\usepackage{enumitem}
\usepackage{booktabs}
\usepackage{graphicx}
\usepackage{makecell}
\usepackage{listings}
\usepackage{float}
\usepackage{fontawesome5}
\usepackage{url}
\urldef{\projecturl}\url{https://lannwei.github.io/RefGuard/} 
\colorlet{rev}{black}
\newcommand{\rev}[1]{{\color{rev}#1}}
\newcommand{\NAE}{\textsc{Nae}}

\title{\LARGE \bf
RefGuard: Identity-Aware Language-Guided Robot Manipulation via Joint Target-Anchor-Frame Grounding
}

\author{Lan Wei$^{1}$, Kangyi Lu$^{1}$, Yongchen Wang$^{1}$, Chenmeng Bi$^{1}$, Qi Chen$^{1}$, \\ Hanlin Niu$^{2,3}$, Yip Fun Yeung$^{4}$, and Dandan Zhang$^{1}$*%
\thanks{$^{1}$Lan Wei, Kangyi Lu, Yongchen Wang, Chenmeng Bi, Qi Chen, and Dandan Zhang are with Imperial College London, UK.
$^{2}$Hanlin Niu is with RACE, United Kingdom Atomic Energy Authority, UK.
$^{3}$Hanlin Niu is also with Oxford Robotics Institute, University of Oxford, UK.
$^{4}$Yip Fun Yeung is with Avant Intelligence Ltd, London, UK}
\thanks{*Corresponding: d.zhang17@imperial.ac.uk}
\thanks{Project page: \projecturl}
}

\begin{document}
\maketitle

\begin{abstract}
Vision-language-action (VLA) models have substantially advanced language-guided robot manipulation, yet reliable execution still hinges on identifying \emph{which} physical object an instruction refers to.
In cluttered scenes containing repeated objects, ambiguous anchors, or frame-dependent spatial terms, a robot can execute a geometrically valid action on a semantically compatible but unintended instance; we call this failure an \emph{identity switch}.
The referent is jointly determined by three coupled latent variables: the target, the anchor, and the reference frame, so committing to any one of them before execution turns residual ambiguity into a silent and irreversible error.
We propose RefGuard, an identity-aware grounding framework that delays commitment by maintaining a joint posterior over all three variables.
RefGuard builds a frame-conditioned object-centric scene graph from RGB-D observations, separating frame-independent geometry from directional relations, and routes the posterior through a decision policy that executes, clarifies, reobserves, or aborts.
On a real UF850 arm, RefGuard records no identity switch on any ambiguity-stress trial and executes correctly on 90.0\% of them, whereas fine-tuned VLA and LLM (Large Language Model)-based baselines switch identity in 33-46\% of the same trials, while retaining 93.3\% success on unambiguous scenes and recovering from post-grounding scene changes in 86.7\% of trials.
On a 3{,}200-episode procedural suite, it raises correct execution on solvable instructions from 56.6\% to 80.5\% over the ablation that commits to the anchor and frame before the target, while deferring less often (19.5\% vs.\ 43.4\%).
\end{abstract}



\section{Introduction}
Language-guided manipulation in cluttered tabletop environments requires grounding each instruction to a unique physical object instance \cite{li2024foundation}. For a command such as ``pick the bell pepper to the right of the can \rev{and place it} into the bowl near the cup'', resolving the referent depends on a chain of coupled choices: which pepper, which can, which bowl, and whether ``right'' is read in the user or robot frame. A system with accurate detection, fluent vision-language reasoning, and reliable grasp planning can execute every step competently yet act on the wrong instance, producing a geometrically valid grasp on an object that matches the description but not the user's intent. Such failures are not attributable to any single module, but emerge from the joint structure of grounding under ambiguity.

\begin{figure}[!t]
\centering
\includegraphics[width=1\linewidth]{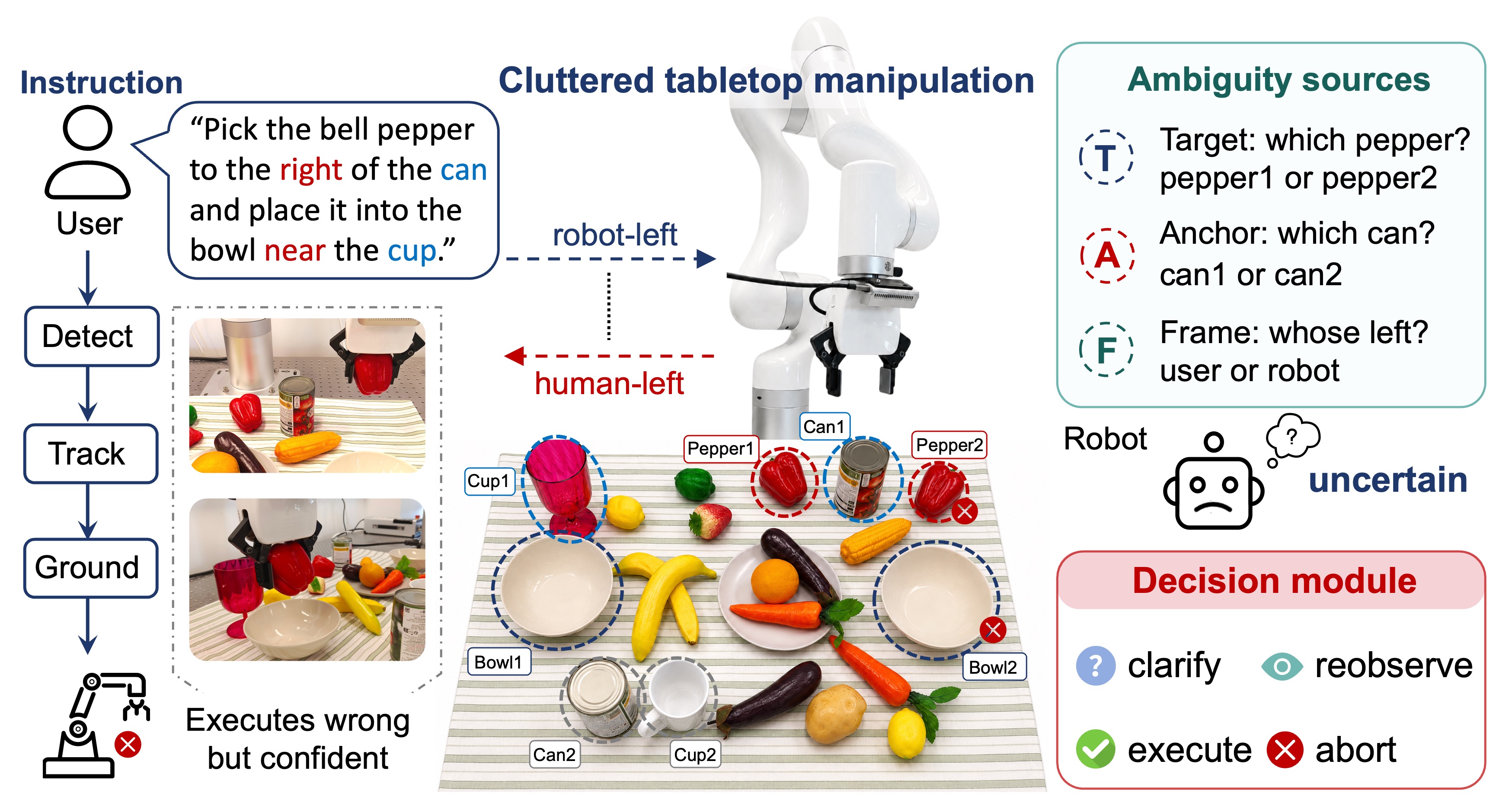}
\vspace{-6mm}
\caption{In cluttered tabletop manipulation, identity switches arise when ambiguity over the target, anchor, or reference frame is collapsed into hard choices before execution. RefGuard models these variables jointly, preserves their uncertainty, and routes the posterior through a decision module that chooses to execute, clarify, reobserve, or abort, reducing confident wrong-object execution.}
\label{fig:1}
\vspace{-7mm}
\end{figure}

We term this failure an \emph{identity switch}, illustrated in Figure~\ref{fig:1}. While referring expression comprehension studies this ambiguity at the language-perception interface \cite{zhou2026roborefer, yuan2024robopoint, you2024ferret}, an identity switch is its downstream consequence at execution time, when a premature grounding commitment is converted into an irreversible physical action on the wrong object. The referent is jointly determined by three coupled latent variables, the target, the anchor, and the reference frame, whose assignments mutually constrain one another. Resolving identity therefore requires modelling this joint uncertainty explicitly and connecting it to a decision of whether the current belief is reliable enough to act on.

\begin{figure*}[!t]
\centering
\includegraphics[width=\textwidth]{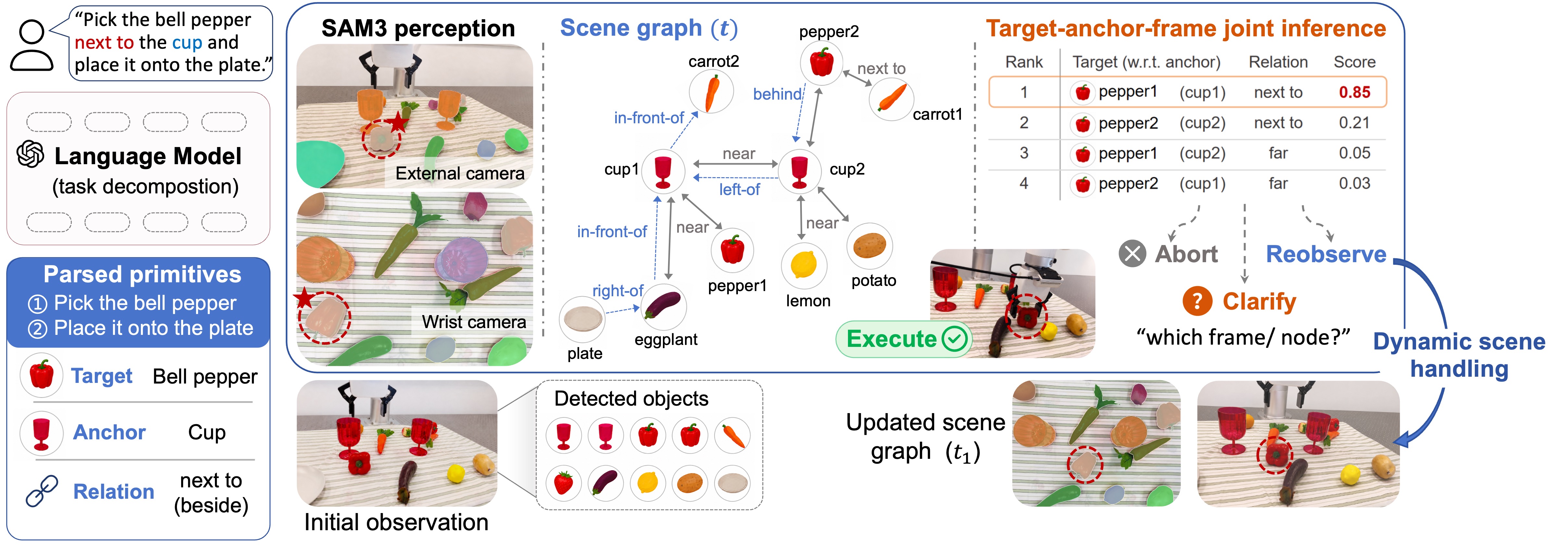}
\vspace{-7mm}
\caption{Overview of RefGuard. A language model decomposes the instruction into symbolic primitive queries over target, anchor, and relation, while SAM~3 perception builds a frame-conditioned scene graph from RGB-D observations. A joint hypothesis table scores candidate target--anchor--frame interpretations, and execution is committed only when the target belief is sufficiently concentrated for a safe action; otherwise the system clarifies, reobserves, or aborts. The scene graph is rebuilt after each primitive to support subsequent ones.}
\vspace{-5mm}
\label{fig:2overview}
\end{figure*}

Existing language-guided manipulation systems often hide this uncertainty. End-to-end VLA policies \cite{zitkovich2023rt,kim2024openvla,wen2025dexvla}, including GR00T \cite{bjorck2025gr00t} and $\pi_{0.5}$~\cite{black2025pi_}, LLM-driven pipelines \cite{ahn2022can, huang2023voxposer, huang2024rekep}, and VLM-based grounders and spatial-reasoning models \cite{zhou2026roborefer, liu2024grounding, bhat2024hifi, cheng2024spatialrgpt} commit early\rev{: they either select a single most likely target, anchor, and frame and pass these hard choices to a downstream planner, or fold the choice implicitly into the predicted actions}. Such \rev{systems} are effective when only one referent is plausible, but in cluttered scenes they collapse residual ambiguity prematurely, \rev{so the action is executed with no record that its referent was uncertain}. One might expect better perception to close this gap, yet tracking-centric systems only preserve assigned identities over time through instance segmentation and identity-preserving trackers \cite{carion2025sam, yang2024samurai}, without determining which identity the language intended in the first place. Others let robots ask for help when uncertain~\cite{ren2023robots, park2023clara}, but choose among plans rather than resolving the coupled target--anchor--frame uncertainty within a single action.

We propose RefGuard, which reformulates identity preservation as joint posterior inference over the target, anchor, and reference frame, replacing commitment to a top-1 referent with late commitment at the grounding-to-action interface.
These variables are coupled, since a different anchor or frame can change which target best fits the instruction even when all objects are correctly detected, making independent estimation unreliable. From RGB-D observations, RefGuard builds a frame-conditioned two-layer scene graph that separates relations by frame dependence for efficient joint inference under ambiguity. A decision policy maps the posterior to action, letting the robot execute when belief is concentrated, query when mass is split, reobserve when evidence is weak, \rev{and abort when no valid interpretation remains}. By deferring commitment until the moment of action, RefGuard equips the robot with awareness of its grounding reliability, distinguishing justified correct execution from confidently mistaken execution.
\rev{RefGuard is complementary to learned LLM/VLM grounders: they can supply language interpretations, ambiguity predictions, or hypothesis scores to this interface, while RefGuard preserves joint uncertainty over the physical target--anchor--frame instances and verifies it before execution, so a fine-tuned grounder is a component rather than an alternative.}
\rev{On a real UF850 arm, RefGuard executes correctly on 81 of 90 ambiguity-stress trials without a single identity switch, and on the solvable episodes of a 3{,}200-episode procedural suite it raises correct execution from 56.6\% to 80.5\% over a matched ablation that commits to the anchor and frame before the target.}

The contributions of this work are as follows:
\begin{enumerate}[itemsep=1pt, parsep=1pt, topsep=1pt]
    \item We characterise identity switches as a failure mode in language-guided robot manipulation, where a robot executes a geometrically valid action on a semantically compatible but physically unintended object. We formulate the underlying ambiguity as joint inference over three coupled variables: the target instance, the anchor instance, and the reference frame.
    \item We introduce RefGuard, a grounding framework built on a two-layer scene graph that separates frame-independent geometry from directional relations, coupled with joint target--anchor--frame posterior inference that preserves referential uncertainty until the decision stage, letting the robot execute when the posterior is reliable, clarify when grounding is ambiguous, reobserve when evidence is weak, or abort when unsupported.
    \item \rev{We evaluate RefGuard on a six-setting UF850 stress suite for identity switches, together with matched early-commitment and scene-update controls, sequential composition, held-out object categories, and a 3{,}200-episode procedural suite. Where VLA and LLM-based baselines execute valid grasps on the wrong instance, RefGuard acts correctly or explicitly defers, and the controls attribute this benefit to preserving joint uncertainty rather than to abstention or scene-update access alone.}
\end{enumerate}

\section{Method}

\subsection{System Overview and Problem Definition}
\label{sec:overview}
The robot receives a natural-language instruction $x$ and calibrated RGB-D observations of the tabletop scene.
In our setup, the scene is observed by an external third-person camera and a wrist-mounted camera.
The perception module\rev{, built on SAM~3~\cite{carion2025sam},} fuses these observations and returns a set of detected object instances $V=\{o_i\}_{i=1}^{N}$.
The language module parses the instruction into an ordered sequence of primitive queries, $\Pi = (Q_1,\ldots,Q_L)$, which are grounded and executed sequentially (Figure~\ref{fig:2overview}).

For each primitive query, RefGuard returns a task contract
$
    \mathcal{T} = (\alpha, \hat{t}, \mathcal{H}_{\hat{t}}, \mathcal{C}, \kappa, d, r_d).
$
Here $\alpha$ is the primitive action, such as ``pick'' or ``place''; $\hat{t}$ is the selected target instance when execution is allowed; $\mathcal{H}_{\hat{t}}$ is the target-conditional hypothesis table retained for planning; $\mathcal{C}$ contains spatial and action-feasibility constraints; $\kappa$ is a posterior-concentration score used by the decision policy; and $r_d$ is a reason code explaining the decision.
The decision belongs to
\[
    d \in \{\textsc{execute}, \textsc{clarify}, \textsc{reobserve}, \textsc{abort}\}.
\]
The decision policy maps the posterior grounding state and feasibility evidence to one of four decisions.
RefGuard executes when the posterior concentrates on a single feasible grounding interpretation and the planner can produce a safe action.
It clarifies when the top hypotheses disagree on the target identity, anchor assignment, or reference-frame interpretation.
It reobserves when perceptual evidence is weak or stale, and aborts when no compatible target, anchor, or satisfiable action constraint exists.
When $d \neq \textsc{execute}$, the target $\hat{t}$ is left uninstantiated and the system queries the user, reobserves the scene, or aborts.
The reason code $r_d$, such as \textsc{confident-target}, \textsc{target-ambiguous}, \rev{or} \textsc{missing-target}, supports both system behaviour and analysis, preventing the planner from silently receiving an overconfident but underspecified target.

Given a primitive query $Q_\ell$, RefGuard seeks a task-relevant interpretation rather than a single object label. An interpretation is unsafe when multiple plausible assignments over target, anchor, or frame induce different executable goals. When all plausible assignments share the same physical goal they are action-equivalent and remain safe. If the language and observations do not determine a unique safe executable goal, the correct behaviour is to return \textsc{clarify}, \textsc{reobserve}, or \textsc{abort} rather than executing an arbitrary top-ranked interpretation.

\subsection{Language Parsing into Primitive Queries}
\label{sec:parsing}
We use OpenAI GPT-4o as a language parser for subtask decomposition \cite{hurst2024gpt, liang2023code}.
Given a natural-language instruction $x$, GPT-4o converts the instruction into an ordered sequence of symbolic primitive queries, $\Pi = (Q_1,\ldots,Q_L)$, which are grounded and executed sequentially by RefGuard.
The parser is constrained to produce only a symbolic task structure and does not act as an instance-level grounder. It neither binds language phrases to detected object instances nor selects a target, anchor, or reference frame, leaving all instance-level assignments to the grounding module.

Each primitive query is represented as
$
    Q_\ell =
    \left(
    \alpha_\ell,\;
    q_{t,\ell},\;
    \{(q_{a,\ell}^{(k)}, r_\ell^{(k)})\}_{k=1}^{m_\ell},\;
    \mathcal{R}^{0}_{\ell},\;
    c_{F,\ell},\;
    \mathcal{C}_{\mathrm{act},\ell}
    \right),
$
where $\alpha_\ell$ is the action (\textsc{pick} or \textsc{place}) and $q_{t,\ell}$ the target phrase. Each pair $(q^{(k)}_{a,\ell}, r^{(k)}_\ell)$ gives an anchor phrase with its relation to the target, allowing several anchors. \rev{$\mathcal{R}^{0}_{\ell}$} holds anchor-free but frame-dependent constraints such as ``the object on my left''. $c_{F,\ell}$ records the reference-frame cue, e.g.\ \textit{robot perspective}, \textit{human view}, or \textit{unspecified}. \rev{$\mathcal{C}_{\mathrm{act},\ell}$} lists action constraints, such as target graspability, placeability, surface support, or containment. For instance, ``place the bell pepper into the bowl near the cup'' parses into $\alpha=\textsc{place}$, $q_t=$ ``bell pepper'', two anchor pairs (``bowl'', \textit{into}) and (``cup'', \textit{near}), empty \rev{$\mathcal{R}^{0}$}, unspecified $c_F$, and a containment constraint in \rev{$\mathcal{C}_{\mathrm{act}}$}.

The cue $c_{F,\ell}$ only records which reference frame the instruction suggests and does not resolve it to a concrete frame. An explicit phrase such as ``from the robot's perspective'' sets $c_{F,\ell}$ to that frame, while an underspecified term such as ``left'' leaves it unspecified so that multiple candidate frames remain active. Constructing the frame prior $\pi_F(f)$ from $c_{F,\ell}$ and scene availability, and resolving any residual frame ambiguity, are deferred to the grounding module.

\subsection{Object-Centric Scene Graph and Candidate Evidence}
\label{sec:scene_graph}
From fused RGB-D observations, each detected instance is a node $o_i = (m_i, u_i, x_i, b_i, c_i, h_i, \sigma_i, \nu_i)$ grouping geometry (mask $m_i$, image \rev{centre} $u_i$, 3D position $x_i$ in the robot base frame, bounding box $b_i$), semantics (category and attribute evidence $c_i$, grasp and affordance metadata $h_i$), and uncertainty (detection uncertainty $\sigma_i$, visibility $\nu_i$), where the latter two drive the perception-weak decision. Node identities are local to the current scene rather than persistent tracking IDs, so the graph is rebuilt and identity re-inferred after each observation, letting RefGuard recover when objects move or are swapped before execution.

RefGuard builds a two-layer scene graph $G = (V, E_{\text{pos}}, \{E_{\text{dir}}^f\}_{f \in F})$ that separates frame-independent geometry from frame-dependent direction. The frame-independent layer $E_{\text{pos}}$ stores pairwise geometry such as metric distance, height difference, bounding-box overlap, contact, and containment, supporting topological relations like \textit{near}, \textit{on}, and \textit{inside}. The frame-conditioned layer $E_{\text{dir}}^f$ stores directional offsets under each candidate frame, allowing relations such as \textit{left-of}, \textit{right-of}, and \textit{in-front-of} without committing to a single frame in advance, and offsets are computed only for frames relevant to the current query.

Each interpretation is scored by three types of evidence. For an object $o_i$ and phrase $q$, the soft object score $E_{\text{obj}}(o_i, q)$ aggregates category match, available attribute match, detector confidence, and visibility, with missing attributes such as colour or size omitted rather than penalised so that absent attributes do not discard correct candidates. The soft relation score $E_{r^{(k)}}(i, j_k; f)$ for the $k$-th anchor phrase uses $E_{\text{pos}}$ for topological relations and $E_{\text{dir}}^f$ for directional ones, leaving weak or near-tie evidence uncertain to later trigger clarification or reobservation.
Finally, action feasibility evaluates graspability, collision margin, visibility, and reachability for picking, and support height, clearance, containment, and relation consistency for placement. This score feeds the row score in the joint hypothesis table, where hard feasibility violations are pruned before normalisation.

\subsection{Joint Hypothesis Table for Target--Anchor--Frame Grounding}
\label{sec:hypothesis_table}
RefGuard grounds each query by jointly inferring the target, anchor, and reference frame, maintained in a hypothesis table. For a primitive query $Q$, each row $h$ represents one complete interpretation,
$
    h = (i, j_1,\ldots,j_m, f, s_h, w_h),
$
where $i$ is the target candidate, $j_k$ the object assigned to the $k$-th anchor phrase, $f$ the reference frame, $s_h$ the row score, and $w_h$ the \rev{normalised} row weight. If the query has no explicit anchor, the anchor tuple is empty and the row retains only the target, frame, and anchor-free constraints.
Each candidate frame $f$ carries a prior $\pi_F(f)$ formed from the parsed cue $c_{F,\ell}$ and its scene availability. An explicit cue concentrates the prior on the requested frame, while an unspecified cue spreads it across the available frames. Unobservable frames, such as a user-centric frame without a known user pose, receive zero availability and are dropped before the table is built. \rev{In our setup the user's pose is not tracked, so a human-view cue is mapped to the external-camera frame, which approximates the user's viewpoint (Section~\ref{sec:main_results}).}
A row is scored by combining target evidence, anchor evidence, frame prior, spatial relation evidence, and action feasibility,
\begin{align*}
    s_h ={}&
    E_{\mathrm{obj}}(o_i,q_t)
    + \sum_{k=1}^{m} E_{\mathrm{obj}}(o_{j_k},q_a^{(k)})
    + \log \pi_F(f) \\
    &+ \sum_{k=1}^{m} E_{r^{(k)}}(i,j_k;f)
    + E_{\mathrm{act}}(h).
\end{align*}
The row weights are obtained by normalising over valid rows:
\[
    w_h = \frac{\exp(s_h)}
    {\sum_{h'\in \mathcal{H}_{\mathrm{valid}}}\exp(s_{h'})}.
\]
\rev{Anchor-free constraints in $\mathcal{R}^{0}$ are evaluated under the same frame $f$ and enter $s_h$ as additional relation terms.} This table-based formulation naturally handles multiple anchors because each row stores one candidate for every anchor phrase. It avoids the failure mode of first committing to an anchor or a frame and then searching for the target.
Rows are rejected before normalisation if they violate hard constraints. Common invalid cases include a target and anchor that must differ but coincide, a required anchor with no compatible candidate, an unavailable required frame, an ungraspable target, a placement target that cannot fit its specified container, and a relation that is geometrically impossible under all available frames.

From the hypothesis table, RefGuard computes a target belief by summing the weights of all rows that select the same target:
$ B_T(i) = \sum\nolimits_{h:\,t(h)=i} w_h . $
Let $\hat{t}=\arg\max_i B_T(i)$ be the top target and $\Delta_T$ the margin between the top two beliefs. RefGuard also forms a top hypothesis set $\mathcal{H}^{\mathrm{top}}$, the smallest set of rows whose cumulative weight exceeds a fixed mass threshold, and diagnoses uncertainty by comparing the variables that differ within it. If the top rows disagree mainly on target identity, the case is labelled \textsc{target-ambiguous}, whereas competing anchor assignments yield \textsc{anchor-ambiguous}. When the rows share target and anchors but differ only in reference frame, RefGuard returns \textsc{frame-ambiguous}, capturing the residual frame uncertainty left by an unspecified cue $c_{F,\ell}$. Low visibility, low detector confidence, or stale poses instead yield \textsc{perception-weak} or \textsc{scene-stale}. This diagnosis determines whether RefGuard executes, clarifies, reobserves, or aborts.

\subsection{Posterior-to-Decision Policy}
\label{sec:decision_policy}
Let $p_1$ and $p_2$ denote the largest and second-largest masses of the target marginal $B_T$, respectively\rev{, so that $\Delta_T=p_1-p_2$, and let $\mathcal{S}_T=\{\,i : B_T(i)>0\,\}$ denote the support of $B_T$}. For $|\rev{\mathcal{S}_T}|=1$, we set $p_2 \coloneqq 0$. Let $H_T = H(B_T)$, and define the \rev{normalised} target entropy as
\[
\bar{H}_T =
\begin{cases}
H_T / \log |\rev{\mathcal{S}_T}|, & |\rev{\mathcal{S}_T}| > 1,\\
0, & |\rev{\mathcal{S}_T}| = 1.
\end{cases}
\]
The posterior-concentration score is
\[
\kappa =
\frac{1}{2}p_1 +
\frac{1}{2}(1-\bar{H}_T).
\]
Thus, for a singleton target set, $p_1=1$ and $\kappa=1$.
The score is used only for decision making and is not treated as a formal calibration metric.

The decision policy applies the following rules:
\vspace{-2mm}
\begin{equation*}
d=
\begin{cases}
\textsc{abort},     & \mathcal{H}_{\mathrm{valid}}=\varnothing,\\[2pt]
\textsc{reobserve}, & p_1<0.50 \lor \kappa<0.40,\\[2pt]
\textsc{clarify},   & p_1-p_2<0.15 \lor H_T>1.50 \lor \NAE,\\[2pt]
\textsc{execute},   & \text{otherwise},
\end{cases}
\end{equation*}
where \NAE{} denotes that the \rev{anchor/frame hypotheses in $\mathcal{H}^{\mathrm{top}}$ (Section~\ref{sec:hypothesis_table})} are not action-equivalent.
Hence, competing anchor or frame hypotheses trigger clarification when they imply different physical goals, even if the target marginal itself is concentrated. Thresholds are selected on 30 development episodes\rev{, disjoint from every evaluation set,} and then frozen for evaluation.
\rev{These rules act on the posterior; in addition, a \textsc{perception-weak} or \textsc{scene-stale} diagnosis (Section~\ref{sec:hypothesis_table}) routes the query to \textsc{reobserve} regardless of $B_T$, and a failed pre-execution check (Section~\ref{sec:planning}) re-routes it before any motion.}

After \textsc{clarify}, the reason code identifies whether the ambiguity concerns the target, anchor, or reference frame. The user's reply is parsed as an additional constraint, inconsistent hypotheses are removed, and the same primitive is re-grounded. Execution resumes if the updated decision is \textsc{execute}, so clarification need not terminate the original task.

\subsection{Posterior-Aware Planning and Verification}
\label{sec:planning}
\begin{figure*}[!t]
\centering
\includegraphics[width=0.9\textwidth]{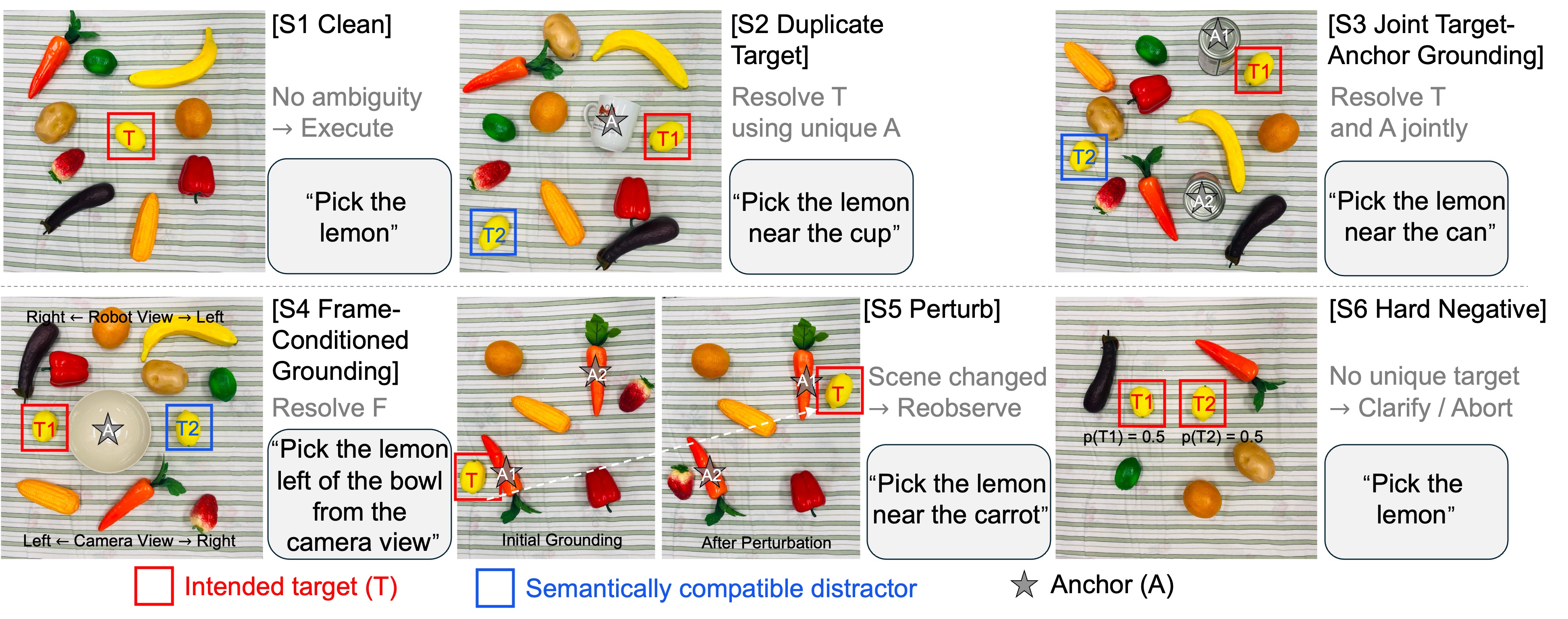}
\vspace{-5mm}
\caption{Identity-switch stress suite. S1 tests ordinary unambiguous execution. S2 tests duplicate targets with a unique anchor. S3 contains duplicate targets and duplicate anchors, requiring joint target--anchor reasoning; geometrically near-tied assignments trigger \textsc{clarify}. S4 tests explicit frame-conditioned grounding. S5 tests robustness under execution-time perturbation. S6 is an ambiguous hard negative, where the correct behaviour is to defer rather than execute an arbitrary action.}
\vspace{-5mm}
\label{fig:stress_suite}
\end{figure*}

\textbf{Posterior-aware planning.} When RefGuard executes, the target is already committed by the belief $B_T$, while the planner retains the high-weight anchor and frame hypotheses so that their coupling is preserved rather than collapsed. For picking, the posterior acts only through this target choice, and the planner samples grasps on the selected object and scores them by graspability, collision margin, reachability, and alignment \cite{fang2020graspnet}.
For placement, it samples poses around the plausible anchors and checks each against the retained spatial constraints \cite{shridhar2022cliport}. A pose is accepted only if it is feasible and consistent with every high-weight hypothesis. When \rev{pose-level checking reveals that} high-weight anchors or frames imply conflicting goals \rev{that the action-equivalence test did not detect}, the query is routed back to \textsc{clarify} or \textsc{reobserve} rather than executed.

\textbf{Pre-execution verification.}
Before motion, a contract check verifies that the target and relevant anchors remain visible, their poses match the grounding result, the spatial constraints still hold, and the action is still feasible.
On failure the robot does not move, routing to \textsc{reobserve} for weak or stale perception, \textsc{clarify} for semantic ambiguity, and \textsc{abort} when no constraint can be satisfied.

\textbf{Execution-time scene update.} After each primitive, the robot reobserves and rebuilds the scene graph. Picked objects are held as in-gripper memory nodes and placed objects are matched near their expected pose, while temporarily occluded objects are kept as memory-backed nodes with reduced confidence. Such memory only bridges successive observations and is overridden once the object is re-detected, so stale memory is never treated as certain evidence. The next primitive is then re-grounded against the updated graph and held/placed-object state; RefGuard does not permanently lock the initial target--anchor--frame binding across a sequence.

\section{Experiments}
We evaluate whether identity-aware grounding reduces confident wrong-object executions in cluttered tabletop manipulation.
The experiments are designed to separate low-level manipulation ability from referential grounding: the first setting contains no identity ambiguity, while the remaining settings progressively introduce duplicate targets, joint target--anchor grounding with duplicate targets and anchors, explicit frame-conditioned grounding, execution-time perturbations, and irreducibly ambiguous instructions.
\rev{Section~\ref{sec:main_results} reports the main real-robot comparison against VLA and LLM-based baselines, Section~\ref{sec:controls} isolates the contribution of joint inference with matched controls, and Sections~\ref{sec:sequential} and~\ref{sec:procedural} extend the evaluation to sequential composition, generalisation, and procedurally generated clutter at scale.}

\subsection{Real-Robot Setup and Baselines}
\label{sec:setup}
We evaluate all methods on a UF850 real-robot tabletop manipulation platform arranged as a kitchen scene.
The robot observes the workspace with two calibrated Intel RealSense D435i RGB-D cameras: an external third-person camera and a wrist-mounted camera.
Each episode starts from a reset tabletop layout and a natural-language instruction.
Additional setup details are in Appendix~\ref{app:uf850_setup}.

We compare RefGuard with three baselines. The first two are VLA policies, GR00T and $\pi_{0.5}$, adapted to our platform. Both are fine-tuned on the same demonstration dataset used for embodiment adaptation, with the same LeRobot-style prompt and action formatting \cite{cadene2026lerobot}.
Each receives the natural-language instruction and robot observations as input and predicts a robot action sequence.
Details are \rev{given in} Appendix~\ref{baseline}.
The LLM-based pipeline uses the same perception and motion planner as RefGuard, but performs early commitment by producing a single hard assignment of the target before planning\rev{; it grounds the scene once at the start of an episode and then executes the resulting plan without re-observation}.

\subsection{Metrics and Identity-Switch Stress Suite}
\label{sec:metrics}

\begin{table*}[!t]
\centering
\caption{Main real-robot results across six UF850 settings, with 30 trials per method and setting. Each cell gives the count with the percentage in parentheses. S1 reports clean-scene task success (TS, $C/N$); S2--S4 report the execution-conditional identity-switch rate $\mathrm{ISR}=I/(C+I)$, whose denominator is the number of executed trials and therefore differs across cells (lower is better); S5 reports post-perturbation recovery; and S6 reports the correct defer rate (CDR; higher safe deferral is better). Best column results are \textbf{bolded}.}
\label{tab:main_results}
\vspace{-2mm}
\small
\setlength{\tabcolsep}{4pt}
\begin{tabular*}{\textwidth}{@{\extracolsep{\fill}} l c c c c c c @{}}
\toprule
Method &
\makecell[c]{S1 Clean \\ TS $\uparrow$} &
\makecell[c]{S2 Dup-T \\ ISR $\downarrow$} &
\makecell[c]{S3 Dup-T/A \\ ISR $\downarrow$} &
\makecell[c]{S4 Frame-Cond. \\ ISR $\downarrow$} &
\makecell[c]{S5 Perturb \\ Recovery $\uparrow$} &
\makecell[c]{S6 Hard-Neg. \\ CDR $\uparrow$} \\
\midrule
GR00T
& 25/30 (83.3\%) & 13/25 (52.0\%) & 10/25 (40.0\%) & 7/25 (28.0\%) & 12/30 (40.0\%) & N/A \\
$\pi_{0.5}$
& 27/30 (90.0\%) & 11/26 (42.3\%) & 9/25 (36.0\%) & 16/27 (59.3\%) & 14/30 (46.7\%) & N/A \\
LLM-based
& 27/30 (90.0\%) & 12/29 (41.4\%) & 12/27 (44.4\%) & 17/30 (56.7\%) & 0/30 (0.0\%) & N/A \\
RefGuard (Ours)
& \textbf{28/30 (93.3\%)} & \textbf{0/28 (0.0\%)} & \textbf{0/27 (0.0\%)} & \textbf{0/26 (0.0\%)} & \textbf{26/30 (86.7\%)} & \textbf{30/30 (100.0\%)} \\
\bottomrule
\end{tabular*}
\vspace{-5mm}
\end{table*}

We design a six-part UF850 stress suite to separate ordinary manipulation ability from identity-aware grounding.
Figure~\ref{fig:stress_suite} gives an example of \rev{each} evaluation condition, with \rev{instruction templates and scene layouts} in Appendix~\ref{instruction}.

We use task-level metrics aligned with the six evaluation settings. Let $N$ be the number of episodes, with $C$ denoting correct execution, $I$ an identity switch, $F$ a low-level failure, and $D$ a defer decision (\textsc{clarify}, \textsc{reobserve}, or \textsc{abort}), so that $N=C+I+F+D$. Where space is limited, we abbreviate an identity switch as a \emph{wrong-ID} execution; the two terms are used interchangeably. For the clean setting S1 we report task success, $\mathrm{TS}=C/N$. For S2--S4, the identity-switch rate is execution-conditional,
\[
\mathrm{ISR}=\frac{I}{C+I},
\]
so its denominator is the number of executed trials, $C+I$, rather than $N$; we therefore always report it together with its counts. We additionally use correct execution $\mathrm{CE}=C/N$, execution coverage $\mathrm{Cov}=(C+I+F)/N$, which also counts attempts that fail at the manipulation level, and the episode-level deferral rate $\mathrm{DR}=D/N$. Every deferral is an intervention that costs a user reply or an extra observation, so on solvable trials lower DR is better. For S5, post-perturbation recovery is the fraction of episodes in which the robot still executes the correct action after the perturbation, typically by detecting the change at the pre-execution check, reobserving, and re-grounding; an episode that ends in a deferral is safe but is not counted as recovered. For hard-negative settings such as S6, we instead report the correct defer rate $\mathrm{CDR}=D/N$, where higher safe deferral is better.

\begin{figure}[!t]
\centering
\includegraphics[width=1\linewidth]{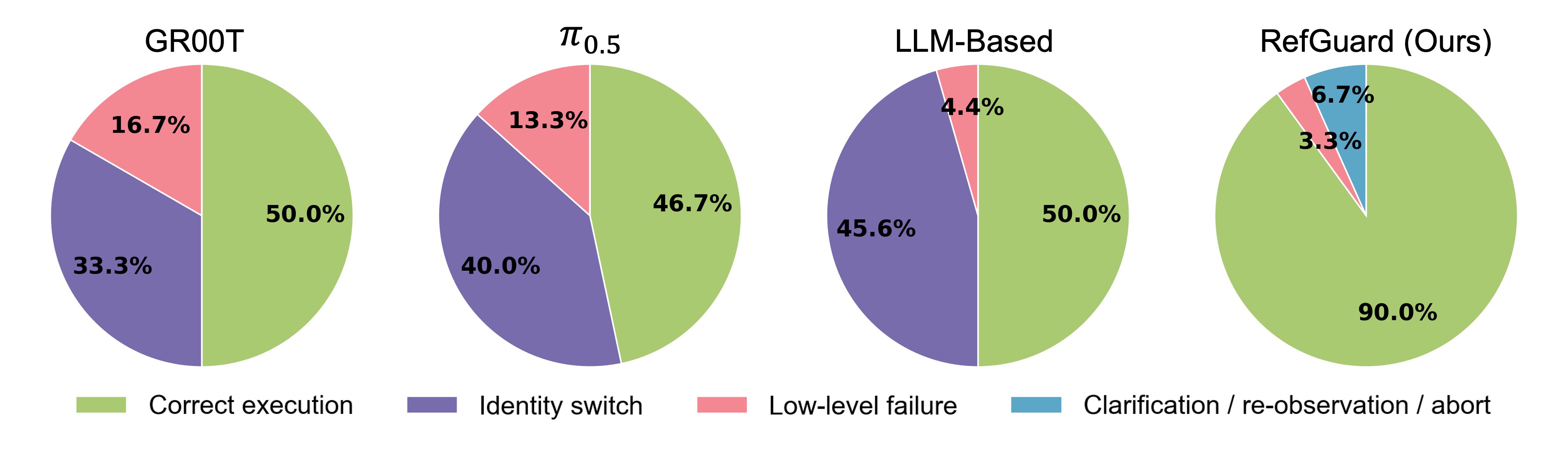}
\vspace{-8mm}
\caption{Outcome decomposition over S2--S4 only ($N=90$ trials per method; no macro-averaging). Episodes are categorised as correct execution/identity switch/low-level failure/defer ($C/I/F/D$): GR00T $45/30/15/0$, $\pi_{0.5}$ $42/36/12/0$, LLM-based $45/41/4/0$, and RefGuard $81/0/3/6$. Percentages in the \rev{figure} are episode-level outcome fractions.}
\vspace{-5mm}
\label{fig:outcome_decomposition}
\end{figure}

\subsection{Main Results}
\label{sec:main_results}
Table~\ref{tab:main_results} reports the main real-robot results across the six
UF850 settings, while Figure~\ref{fig:outcome_decomposition} decomposes
episode-level outcomes over the solvable ambiguity settings S2--S4.

\rev{\textbf{Clean scenes (S1).}}
When the referent is unambiguous, RefGuard achieves the strongest task
success and rarely defers. Thus, its safety mechanism does not come at the
cost of ordinary execution when grounding is clear.

\rev{\textbf{Identity-aware grounding (S2--S4).}}
The main advantage appears when multiple physically plausible referents are
present. Across duplicate targets (S2), duplicate targets and anchors (S3),
and frame-conditioned instructions (S4), the baselines frequently execute
valid motions on the wrong physical instance, whereas RefGuard produces no
observed identity switches. In S3, the anchor is not identifiable in
isolation and must be resolved jointly with the target through the spatial
relation, directly testing the proposed joint formulation. In S4, the
instruction explicitly specifies the reference frame, while the calibrated
robot and external-camera frames imply different target instances; phrases
such as ``my left'' use the external-camera frame as an approximation of the
user viewpoint. Instructions without a resolvable frame instead belong to
S6. These results show that semantic or geometric plausibility alone is
insufficient when instance identity depends on the target--anchor--frame
combination. A separate duplicate-container \textsc{place} variant gives the same
conclusion for receptacle selection: RefGuard uses the secondary spatial
relation to identify the intended container rather than committing to an
arbitrary duplicate.

\rev{\textbf{Scene changes and hard negatives (S5--S6).}}
Under post-grounding perturbations in S5, RefGuard achieves substantially
stronger recovery by reobserving and verifying the grounding before
execution. The original LLM pipeline instead executes a binding formed
before the scene change; because this also differs in access to post-change
observations, Table~\ref{tab:controlled} provides shared-update controls to
isolate the contribution of joint grounding. In S6, where the available
evidence does not support a unique safe action, RefGuard consistently
defers rather than committing arbitrarily. Baselines without an explicit
defer or abort interface are therefore reported as N/A for this setting.

\rev{\textbf{Overall.}}
Figure~\ref{fig:outcome_decomposition} shows that the improvement over
S2--S4 is not obtained by simply replacing executions with deferrals:
RefGuard retains substantially more correct executions while eliminating
observed wrong-ID outcomes. Together, the results indicate that preserving
target--anchor--frame identity through grounding, verification, and
execution improves safety without sacrificing useful execution coverage.

\subsection{\rev{Ablation and Shared-Update Controls}}
\label{sec:controls}

\begin{table}[!t]
\centering
\caption{Matched \rev{real-robot} ablation, 10 trials per method and setting (30 per block).
Entries are $C/I/F/D$: correct execution / identity switch / low-level failure /
defer. $\mathrm{Cov}=(C+I+F)/N$, $\mathrm{CE}=C/N$, and $\mathrm{ISR}=I/(C+I)$.}
\vspace{-2mm}
\label{tab:controlled}
\footnotesize
\setlength{\tabcolsep}{2pt}
\begin{tabular}{@{}lcccccc@{}}
\toprule
Method & S2 & S3 & S4 & Cov.$\uparrow$ & CE$\uparrow$ & ISR$\downarrow$ \\
\midrule
Hard-A/F
& $8/1/1/0$ & $4/2/1/3$ & $7/1/1/1$
& $26/30$ & $19/30$ & $4/23$ \\
RefGuard
& $9/0/1/0$ & $8/0/1/1$ & $9/0/0/1$
& $28/30$ & $26/30$ & $0/26$ \\
\midrule
\multicolumn{7}{c}{S5: same post-change observation and graph update} \\
Method & Swap-T & Move-A & Insert-D & Recovery$\uparrow$ & \multicolumn{2}{c}{ISR$\downarrow$} \\
\midrule
LLM+Update
& $8/1/1/0$ & $7/2/1/0$ & $7/2/1/0$
& $22/30$ & \multicolumn{2}{c}{$5/27$} \\
Hard-A/F+Update
& $8/0/1/1$ & $7/2/0/1$ & $8/1/1/0$
& $23/30$ & \multicolumn{2}{c}{$3/26$} \\
RefGuard
& $9/0/1/0$ & $9/0/1/0$ & $8/0/1/1$
& $26/30$ & \multicolumn{2}{c}{$0/26$} \\
\bottomrule
\end{tabular}
\vspace{-0.5cm}
\end{table}

\begin{figure*}[!t]
  \centering
  \includegraphics[width=0.9\textwidth]{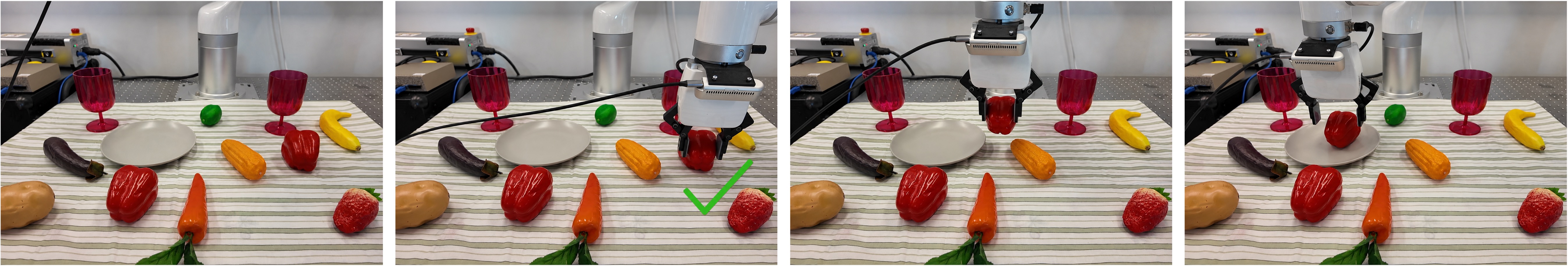}
  \vspace{-4mm}
  \caption{Sequential episode for ``Pick the red pepper near the cup, and place it onto the plate.'' Two red peppers and two cups are present, so only one joint target-anchor assignment satisfies \emph{near}. Left to right: initial scene; RefGuard grasps the pepper adjacent to the right-hand cup; the pepper is held as an in-gripper memory node while the scene graph is rebuilt; the PLACE primitive is executed.}
  \vspace{-2mm}
  \label{fig:sequential}
\end{figure*}

\begin{table*}[!t]
\centering
\caption{Sequential composition, held-out-category generalisation, free-form parsing, and post-action checkpoint results. \rev{Wrong-ID counts identity-switch executions over all primitives.}}
\vspace{-3mm}
\label{tab:sequential}
\small
\setlength{\tabcolsep}{7pt}
\begin{tabular}{@{}lrrrrrr@{}}
\toprule
Condition & Ep. & Prim. & Prim. Comp.$\uparrow$
& Seq. Succ.$\uparrow$ & Wrong- ID$\downarrow$ & Defer$\downarrow$ \\
\midrule
$L=2$ & 10 & 20 & 20/20 (100.0\%) & 10/10 (100.0\%)
& 0/20 & 0/20 \\
$L=4$ & 10 & 40 & 38/40 (95.0\%) & 8/10 (80.0\%)
& 0/40 & 1/40 (2.5\%) \\
$L=6$ & 10 & 60 & 56/60 (93.3\%) & 7/10 (70.0\%)
& 0/60 & 2/60 (3.3\%) \\
\textbf{All} & \textbf{30} & \textbf{120}
& \textbf{114/120 (95.0\%)} & \textbf{25/30 (83.3\%)}
& \textbf{0/120} & \textbf{3/120 (2.5\%)} \\
\rev{Held-out categories (2)} & 18 & 72
& 68/72 (94.4\%) & 15/18 (83.3\%)
& 0/72 & 1/72 (1.4\%) \\
\bottomrule
\end{tabular}

\medskip
\setlength{\tabcolsep}{7pt}
\begin{tabular}{@{}lrrrc@{}}
\toprule
Evaluation & $N$ & Metric & RefGuard & Control \\
\midrule
Free-form paraphrases & 30 & Prog.\ EM$\uparrow$
& 27/30 (90.0\%) & -- \\
Post-action checkpoints & 30 & Dyn.\ Acc.$\uparrow$
& \textbf{28/30 (93.3\%)} & 10/30 (33.3\%) \\
\bottomrule
\end{tabular}
\vspace{-5mm}
\end{table*}

We use matched controls to isolate two possible explanations for RefGuard's
advantage: access to abstention and access to post-change observations.
\rev{Hard-A/F} shares RefGuard's parser, RGB-D observations, scene graph,
evidence scores, four-way decision interface, planner, controller, and
evaluation episodes, but commits to the anchor and reference frame before
inferring the target. Thus, the S2--S4 comparison isolates the effect of
joint target--anchor--frame grounding rather than abstention
(Table~\ref{tab:controlled}, top). RefGuard achieves higher correct execution
while eliminating the identity switches produced by early commitment, with
the largest gap under duplicate-anchor ambiguity (S3). This shows that
prematurely fixing one component can both select the wrong instance and
unnecessarily eliminate otherwise valid hypotheses.

For S5, we additionally give the controls the same post-change observation
and scene-graph update as RefGuard (Table~\ref{tab:controlled}, bottom).
Updating the scene substantially improves both controls, confirming that
re-observation is important, but identity switches remain. RefGuard retains
the strongest recovery performance without observed switches. Therefore,
post-action updating alone is insufficient: reliable recovery also requires
preserving joint grounding uncertainty when re-binding the changed scene.

\subsection{Sequential Composition and Generalisation}
\label{sec:sequential}
We evaluate whether RefGuard can preserve identity-aware grounding over
multi-step instructions containing 2--6 primitives. After each primitive,
the system reobserves the scene, updates held and placed object states, and
re-grounds the next primitive on the updated scene graph
(Section~\ref{sec:planning}).

Figure~\ref{fig:sequential} illustrates this process in a two-primitive
episode with duplicate targets and anchors. The initial PICK is resolved by
the joint target--anchor hypothesis rather than either object independently.
After grasping, the selected pepper is retained as an in-gripper memory node,
and the subsequent PLACE is grounded on the rebuilt scene graph rather than
reusing the initial binding.

As summarised in Table~\ref{tab:sequential}, RefGuard maintains high primitive
completion as sequence length increases, while producing no identity-switch
executions. The same behaviour extends to two held-out object categories,
indicating that the sequential grounding mechanism is not tied to the
development categories. Free-form paraphrase evaluation further shows that
the structured parser generalises beyond the instruction templates.
Most importantly, at post-action checkpoints, RefGuard substantially
outperforms the locked-binding control, supporting dynamic re-grounding after
scene changes rather than permanently retaining the initial
target--anchor--frame assignment.

\subsection{Procedural LIBERO-Style Evaluation at Scale}
\label{sec:procedural}
\begin{figure}[!t]
\centering
\includegraphics[width=1\linewidth]{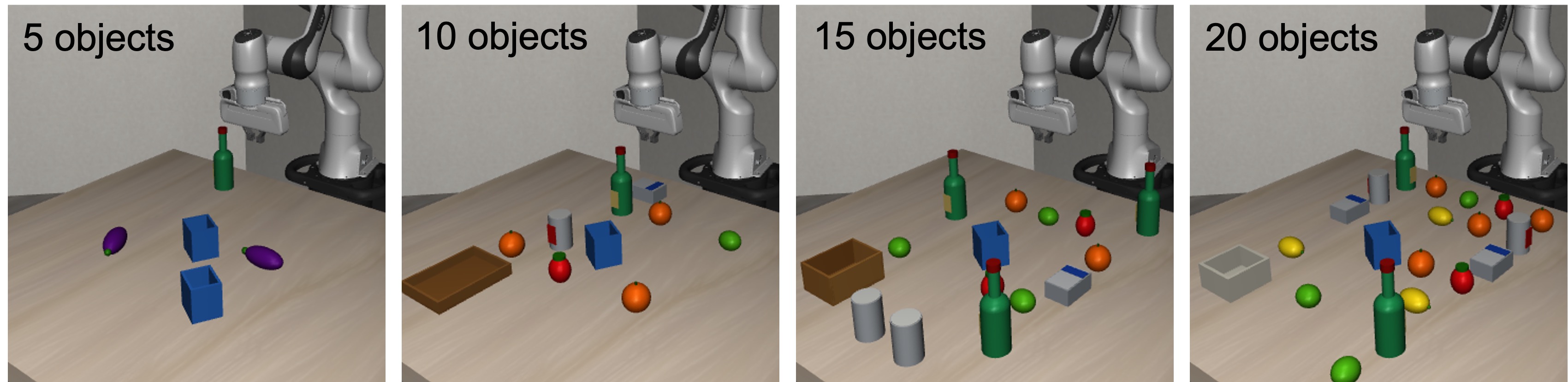}
\vspace{-6mm}
\caption{Example scenes from the procedural LIBERO-style suite as clutter increases from 5 to 20 objects. These scenes are procedurally generated and are not the official LIBERO benchmark.}
\vspace{-3mm}
\label{fig:libero_scale}
\end{figure}

We further test whether RefGuard's advantage persists under substantially
larger and more cluttered scenes using 3{,}200 development-disjoint episodes
from a procedural LIBERO-style suite (Fig.~\ref{fig:libero_scale}), rather
than the official LIBERO benchmark. The suite spans 20 object categories,
5--20 objects per scene, and both solvable and hard-negative instructions.
RefGuard and Hard-A/F receive identical observations and use the same
four-way decision interface, isolating the effect of preserving joint
target--anchor--frame uncertainty.

As shown in Table~\ref{tab:procedural_scale}, RefGuard consistently achieves
higher correct execution than the matched control at every clutter level,
while substantially reducing unnecessary deferral on solvable episodes.
The advantage remains pronounced even in the most cluttered scenes.
Conversely, on hard-negative episodes, RefGuard safely defers in all cases,
whereas Hard-A/F frequently commits despite insufficient evidence.
Neither method exhibits an identity switch among its executed solvable
trials in this evaluation.
Together, these results show that RefGuard's joint uncertainty representation
remains effective as scene complexity increases, improving both execution
coverage when the instruction is resolvable and conservative deferral when
it is not. This procedural study tests scalability under controlled ambiguity
and clutter, rather than unrestricted open-world manipulation.

\begin{table}[!t]
\centering
\caption{Procedural LIBERO-style evaluation over 2{,}400 solvable and 800 hard-negative episodes. CE is correct execution; solvable deferral is lower-better, whereas safe deferral on hard negatives is higher-better.}
\vspace{-3mm}
\label{tab:procedural_scale}
\footnotesize
\setlength{\tabcolsep}{2.5pt}
\begin{tabular}{@{}lccccc@{}}
\toprule
Method & CE@5 & CE@10 & CE@15 & CE@20 & \makecell{Total solvable\\CE$\uparrow$} \\
\midrule
Hard-A/F & $375/600$ & $354/600$ & $323/600$ & $306/600$ & $1358/2400$ (56.6\%) \\
RefGuard & $524/600$ & $495/600$ & $470/600$ & $442/600$ & $1931/2400$ (80.5\%) \\
\bottomrule
\end{tabular}

\medskip
\setlength{\tabcolsep}{4pt}
\begin{tabular}{@{}lccc@{}}
\toprule
Method & ISR$\downarrow$ & \makecell{Solvable\\defer$\downarrow$} & \makecell{Hard-negative\\safe defer$\uparrow$} \\
\midrule
Hard-A/F & $0/1358$ & $1042/2400$ (43.4\%) & $622/800$ (77.8\%) \\
RefGuard & $0/1931$ & $469/2400$ (19.5\%) & $800/800$ (100.0\%) \\
\bottomrule
\end{tabular}
\vspace{-5mm}
\end{table}

\section{Conclusion}
We identify \emph{identity switches} as a key failure mode in language-guided
robot manipulation: a robot may execute a valid action while acting on the
wrong physical referent. RefGuard addresses this failure by preserving joint
uncertainty over the target, anchor, and reference frame until sufficient
evidence supports execution, otherwise triggering clarification,
re-observation, or abort.
Across real-robot experiments, matched controls, sequential tasks, and
large-scale procedural evaluation, RefGuard consistently reduces wrong-ID
execution without degrading performance when the referent is clear. The
controlled comparisons further show that this benefit comes from joint,
delayed commitment rather than abstention or scene-update access alone.
These results suggest that reliable language-guided manipulation requires
more than selecting the most likely grounding: the system must preserve
referential uncertainty through execution and commit only when the intended
physical instance is sufficiently determined.


\bibliographystyle{IEEEtran}
\bibliography{main}

\clearpage
\appendix
\section{Appendix}

\subsection{UF850 Setup}
\label{app:uf850_setup}
The UF850 tabletop set contains 15 produce instances from 10 categories. Five categories appear as duplicate pairs, namely bananas, carrots, bell peppers, lemons, and eggplants, while the remaining five categories appear as singletons. We also include common tableware objects, such as bowls, plates, cups, and containers. All produce and tableware objects can serve as targets or anchors depending on the instruction. This object set supports both ordinary language-conditioned pick-and-place tasks and identity-switch stress tests. Across trials, duplicate or visually similar objects are arranged to create target and anchor ambiguity. These layouts require the robot to resolve the intended referent not only from object category, colour, or shape, but also from relational cues and reference-frame-dependent spatial language.

\subsection{Baseline Implementation Details}
\label{baseline}
\paragraph{$\pi_{0.5}$ baseline.}
We \rev{fine-tune} $\pi_{0.5}$ on our LeRobot-format UF850 manipulation dataset, which contains 300 UF850 demonstrations collected via Meta Quest teleoperation across S1--S4 tasks.
The model \rev{is} initialised from the public \rev{$\pi_{0.5}$} base checkpoint and adapted with LoRA modules in both the PaliGemma backbone and the action expert. We use an action horizon of 15 and disable discrete state inputs.
The LeRobot data pipeline provides the top-view image, wrist image, robot state, action sequence, and task prompt to the model; prompts are read from the dataset task field, images are resized to \rev{$224\times224$}, and an additional delta-action transform is applied to the action sequence. We train for 20,000 gradient steps on 4 NVIDIA A100 80GB GPUs using FSDP over 4 devices, with a global batch size of 64. Optimisation uses AdamW with gradient clipping at 1.0 and a learning rate of \rev{$5\times10^{-5}$} after 1,000 warm-up steps.

At inference time, the policy receives two RGB observations, the 7-dimensional proprioceptive state, and the language instruction. The state consists of six UF850 joint angles and the gripper position, while the images are captured from the main and wrist cameras and resized with padding to \rev{$224\times224$}. Joint commands are sent through UF850 servo control, and the gripper command is clipped to \rev{$[0,1]$} before being mapped to the physical gripper range. This receding-horizon loop is repeated until the episode terminates. Table~\ref{tab:inference_hyperparams} summarises the main inference parameters.

\begin{table}[!t]
\centering
\caption{Inference hyperparameters and notation used during $\pi_{0.5}$ real-robot deployment.}
\label{tab:inference_hyperparams}
\footnotesize
\setlength{\tabcolsep}{4pt}
\begin{tabular}{@{}l l p{0.47\linewidth}@{}}
\toprule
Notation & Shape/Value & Description \\
\midrule
$N_c$ & 2 & number of RGB camera views \\
$X$ & $224 \times 224 \times 3$ & resized RGB observation size \\
$R_{\mathrm{raw}}$ & $640 \times 480$ & raw camera resolution \\
$f_{\mathrm{cam}}$ & 30 Hz & camera frame rate \\
$\mathbf{s}$ & $\mathbb{R}^{7}$ & robot state: 6 joint angles and gripper position \\
$\mathbf{a}$ & $\mathbb{R}^{7}$ & action: 6 joint commands and gripper command \\
$f_{\mathrm{ctrl}}$ & 20 Hz & low-level control frequency \\
$v_{\mathrm{servo}}$ & 120 & servo speed used by the robot API \\
$a_{\mathrm{servo}}$ & 2000 & servo acceleration used by the robot API \\
\bottomrule
\end{tabular}
\end{table}

\paragraph{\rev{GR00T baseline.}}
We also \rev{fine-tune} NVIDIA GR00T-N1.7-3B as \rev{a second VLA} baseline using the same LeRobot-format UF850 manipulation dataset.
The model \rev{is} initialised from the public base checkpoint\rev{ and fine-tuned} on 4 NVIDIA A100 GPUs with a global batch size of 128 for 20,000 optimisation steps\rev{, using} a learning rate of \rev{$1\times10^{-4}$}, \rev{a} weight decay of \rev{$1\times10^{-5}$}, \rev{and} a warm-up ratio of 0.05.
We also \rev{use} the default image augmentation \rev{of} the fine-tuning script, including colour jitter over brightness, contrast, saturation, and hue.

For deployment, at each control cycle, the client reads the current robot state, consisting of six joint angles and the gripper position, and captures RGB observations from the top-view and wrist cameras. The observation is packed into GR00T's modality format with two video streams, the single-arm state, the gripper state, and the language instruction.
The server returns decoded physical actions, represented as an action chunk containing six absolute joint targets and one gripper command per step. By default, we enable temporal action ensembling: each newly predicted chunk is added to a short history, and the executed action is computed as an exponentially weighted average of valid predictions. The resulting joint target is sent to the UF850 through servo control, while the gripper command is mapped to the physical gripper range.

\paragraph{\rev{LLM-based pipeline.}}
\rev{The LLM-based pipeline shares RefGuard's perception stack and motion planner. It grounds the instruction once at the start of an episode by committing to a single hard target (and, where required, anchor) assignment, and then executes the resulting plan without re-observation, exposing no explicit defer or abort decision.}

\paragraph{Matched modular controls.}
Hard-A/F shares RefGuard's language parser, RGB-D observations, object-centric graph, candidate and evidence scores, four-way decision interface, planner, controller, and evaluation episodes. It differs by committing the anchor and reference frame before target inference rather than retaining their joint uncertainty with the target. In S5, LLM+Update and Hard-A/F+Update receive the same post-change observation and graph update as RefGuard; the former otherwise retains the LLM pipeline's hard target assignment, while the latter retains Hard-A/F's early anchor/frame commitment.

\subsection{Sequential and Generalisation Evaluation}
\label{app:sequential_setup}
The sequential-composition evaluation contains 10 episodes each at lengths $L=2$, $4$, and $6$, for 120 primitives in total. Each primitive boundary invokes a fresh observation, object-state update, scene-graph update, and re-grounding of the next primitive. The object-generalisation split contains two additional evaluation-only categories held out from the corresponding development setup, with 18 episodes and 72 primitives. Separately, 30 free-form paraphrases test program exact match under the fixed structured-output parsing schema, and 30 post-action checkpoints compare dynamic re-grounding accuracy with a locked-binding control. Program exact match evaluates parsing and is not treated as physical task success.

\subsection{Instruction Templates and Scene Layouts}
\label{instruction}
We group the instruction templates and scene layouts into six evaluation settings.
Bracketed placeholders are instantiated with object categories, anchors, containers, or reference frames from the current scene.
S1 is an unambiguous sanity check.
S2--S4 test duplicate-target grounding, joint target--anchor grounding with duplicate targets and anchors, and explicit frame-conditioned grounding.
S5 evaluates robustness under post-grounding perturbation.
S6 contains hard negatives in which the instruction and scene do not specify a unique action.

\vspace{0.3cm}

\begin{lstlisting}[frame=tblr, basicstyle=\footnotesize\ttfamily]
S1 Clean
  Templates:
    - "Pick the [object]."
    - "Place the [object] next to the
      [unique anchor]."
  Layout:
    - The target or anchor is uniquely
      identifiable.
    - No duplicate target, duplicate anchor,
      or frame ambiguity is present.

S2 Duplicate Target
  Templates:
    - "Pick the [target] next to the
      [unique anchor]."
    - "Pick the [target] closest to the
      [unique anchor]."
  Layout:
    - Multiple same-category target
      candidates are present.
    - A unique anchor and a spatial relation
      disambiguate the intended target.

S3 Joint Target-Anchor Grounding
  Templates:
    - "Pick the [target] next to the [anchor]."
    - "Place the [target] into the [container]
      near the [secondary anchor]."
  Layout:
    - Duplicate targets and duplicate anchors
      are present; anchors are not
      independently identifiable in isolation.
    - Exactly one joint target-anchor
      assignment satisfies the RGB-D next-to
      relation.
    - Geometrically near-tied assignments
      trigger CLARIFY.
    - In the duplicate-container PLACE variant
      (reported separately in the main text),
      a secondary relation uniquely identifies
      the intended receptacle.

S4 Frame-Conditioned Grounding
  Templates:
    - "Pick the [target] left of the [anchor]
      from the robot's perspective."
    - "Pick the [target] on my left of the
      [anchor]." [external-camera frame]
    - "Place the [target] to the right of the
      [anchor] from the camera view."
  Layout:
    - Every instruction contains an anchor
      phrase and a directional relation
      (left-of / right-of / in-front-of)
      together with an explicit
      reference-frame cue.
    - Duplicate targets are present. The
      anchor category may also be duplicated,
      so the anchor is not identifiable in
      isolation and must be resolved jointly
      with the target and the named frame.
    - Calibrated robot and external-camera
      frames imply different target instances
      for the same anchor and relation, so a
      system that ignores the frame cue, or
      that fixes the anchor before evaluating
      the frame-conditioned relation, acts on
      the wrong instance.
    - Unavailable or stale frame observations
      trigger REOBSERVE.

S5 Perturbation
  Interventions:
    - Target swapped after grounding.
    - Anchor moved after grounding.
    - Distractor inserted between grounding
      and execution.
  Layout:
    - The initial grounding may become stale
      before execution.
    - The robot must verify the scene contract
      and reobserve instead of executing an
      invalidated action.

S6 Hard Negatives
  Conditions:
    - Duplicate identical targets with no
      relational disambiguator.
    - Duplicate anchors producing equally
      valid targets.
    - Symmetric scene where left/right depends
      on an unspecified frame.
    - Instruction with a missing target or
      missing anchor.
  Expected behaviour:
    - No unique safe physical goal is
      determined by the instruction and scene.
    - The correct behaviour is to clarify,
      reobserve, or abort rather than execute
      an arbitrary valid action.
\end{lstlisting}

\end{document}